\documentclass{article}
\usepackage{rsj2018e_IS}
\pdfoutput=1
\usepackage{url}

\usepackage{amsmath}
\usepackage{graphics} % for pdf, bitmapped graphics files
\usepackage{graphicx} % for pdf, bitmapped graphics files

\usepackage{amssymb}  % assumes amsmath package installed
\usepackage{bm}
\usepackage{subfig}

\begin{document}

\title{
Deep Learning with Predictive Control for Human Motion Tracking
}

\author{
Don Joven Agravante\thanks{corresponding author: don.joven.r.agravante@ibm.com}
~
Giovanni De Magistris,
Asim Munawar,
\\
Phongtharin Vinayavekhin,
and
Ryuki Tachibana
\\
IBM Research - Tokyo,  IBM  Japan, Japan.
}

\engtitle{}
\engauthor{}

\setlength{\baselineskip}{4.4mm}
\maketitle
\thispagestyle{empty}
\pagestyle{empty}

\maketitle

%%% Custom Commands
\newcommand{\M}[1]{\mathbf{#1}}
\newcommand{\V}[1]{\mathbf{#1}}
\newcommand{\jointpos}{\V{q}}
\newcommand{\norm}[1]{\left\lVert#1\right\rVert}
\newcommand{\timestep}{\Delta t}

% NOTE: abstract is not needed for RSJ format
\begin{abstract}

\textit{\textbf{Abstract - }} We propose to combine model 
predictive control with deep learning for the task of 
accurate human motion tracking with a robot. We design the MPC to allow switching 
between the learned and a conservative prediction. We also explored online 
learning 
with a DyBM model. We applied this method to human handwriting motion tracking with 
a 
UR-5 robot. The results show that the framework significantly improves tracking 
performance.
\end{abstract}

%%%
\section{Introduction}
Accurate control for human motion tracking is a key requirement in many 
applications 
including human-robot interaction, teleoperation systems, exoskeletons and 
surveillance systems.
For these applications, better motion prediction models enable better control.
But as the system complexity increases, conventional methods which require 
handcrafted models also become increasingly challenging to design.
Deep learning architectures can provide this model given enough data. 

%-- short SoA
Using deep neural networks \textit{end-to-end} for difficult robot 
manipulation tasks was proposed in~\cite{levine:jmlr:2016} but it is too
data-inefficient. Contrary to this, several recent approaches have used a 
Model Predictive Control (MPC) framework such that deep 
learning is used only in the part which is difficult to model.
For example,~\cite{lenz:rss:2015} learns complex contact dynamics for robotic 
food-cutting. In~\cite{finn:icra:2017}, the mapping of actions to image 
pixel motion is learned for vision-based manipulation tasks.
In~\cite{erickson:arxiv:2017}, a model is learned for predicting forces in a 
robot-assisted dressing task. In~\cite{williams:icra:2017}, the dynamics of 
aggressive driving is learned for controlling an autonomous car.

%-- human motion
This work is in the same area of research where neural networks learn complex 
predictive models for use within an MPC framework. Specifically, we 
are learning models to predict human motion.
As the representative task, the robot here has to write characters at the same time 
as a human, as shown in Fig.~\ref{fig: rviz simulations top view}. We chose this 
task to leverage existing datasets on character writing such 
as~\cite{laviola:pami:2007}. In addition to being a new application area, we also 
design the MPC to be able to switch to a more conservative prediction. Furthermore, 
we 
also explore online learning using the Dynamic 
Boltzmann Machine~\cite{osogami:naturesr:2015} neural network.

\newcommand{\lettertrim}[1]{\includegraphics[trim={6cm 0.1cm 0.5cm 1cm}, clip, 
width=0.32\columnwidth]{#1}}
\begin{figure}[!ht]
  \centering
  \subfloat{\lettertrim{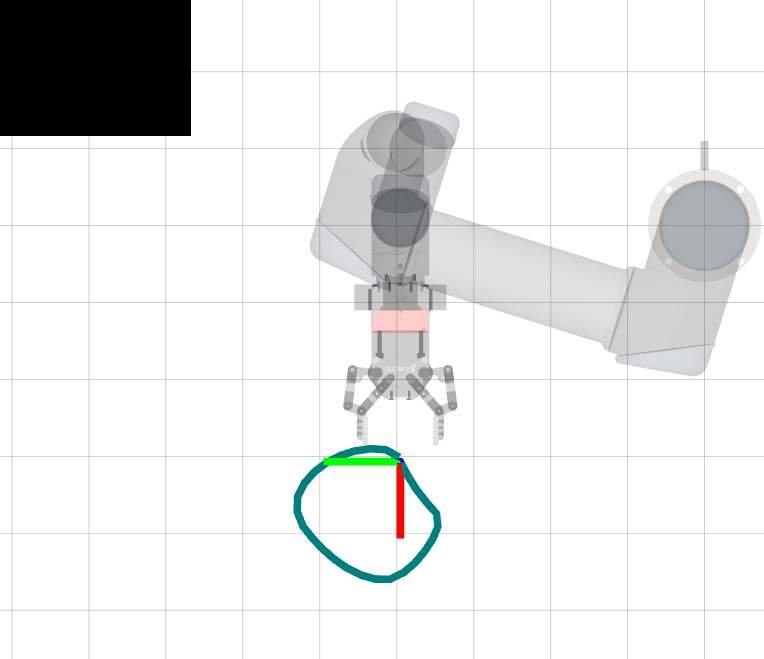}}
  \subfloat{\lettertrim{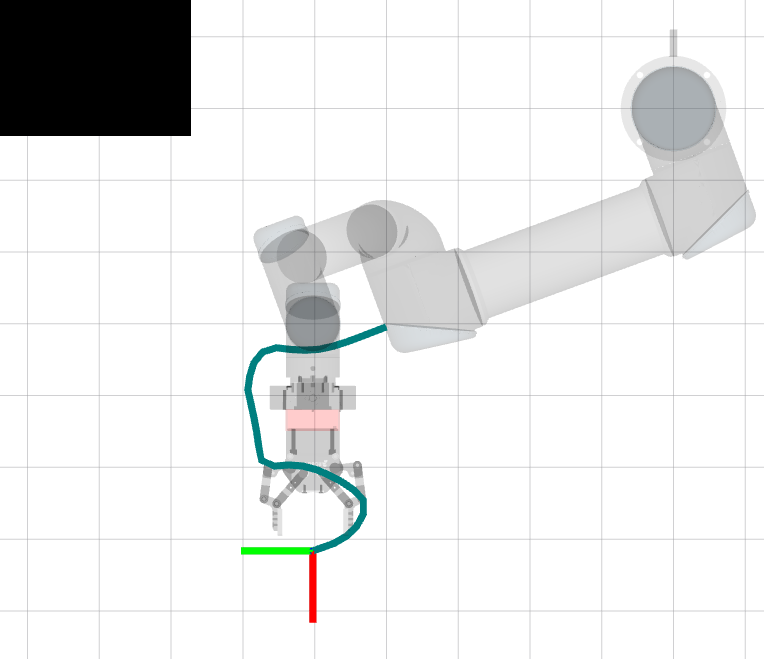}}
  \subfloat{\lettertrim{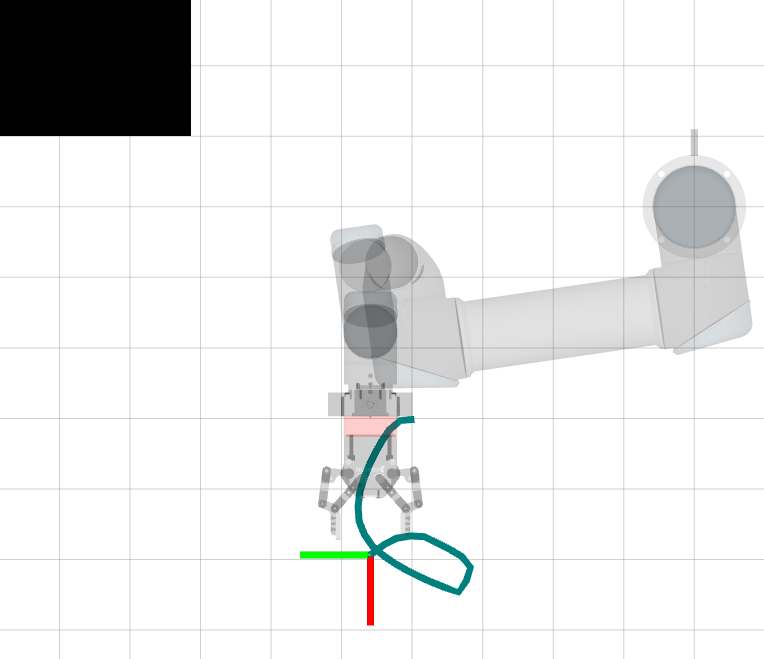}}
 
  \subfloat{\lettertrim{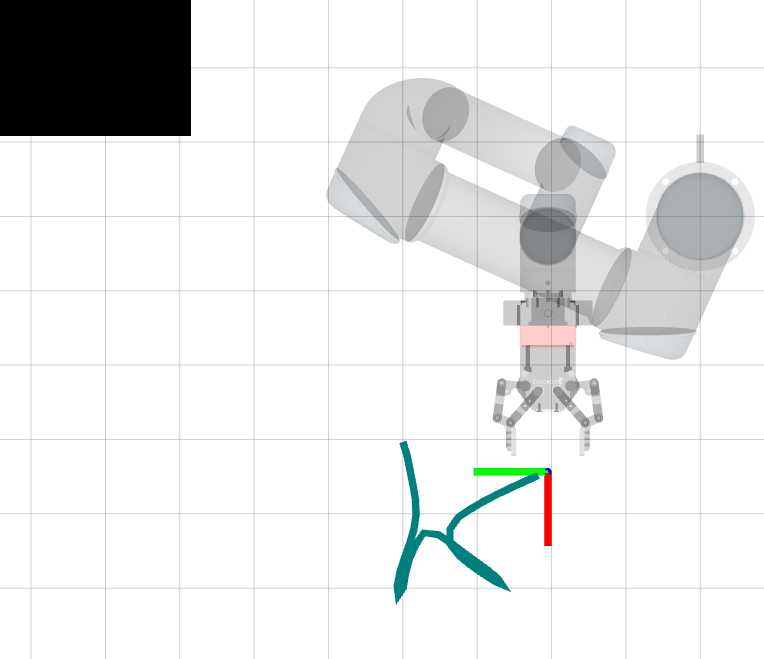}}
  \subfloat{\lettertrim{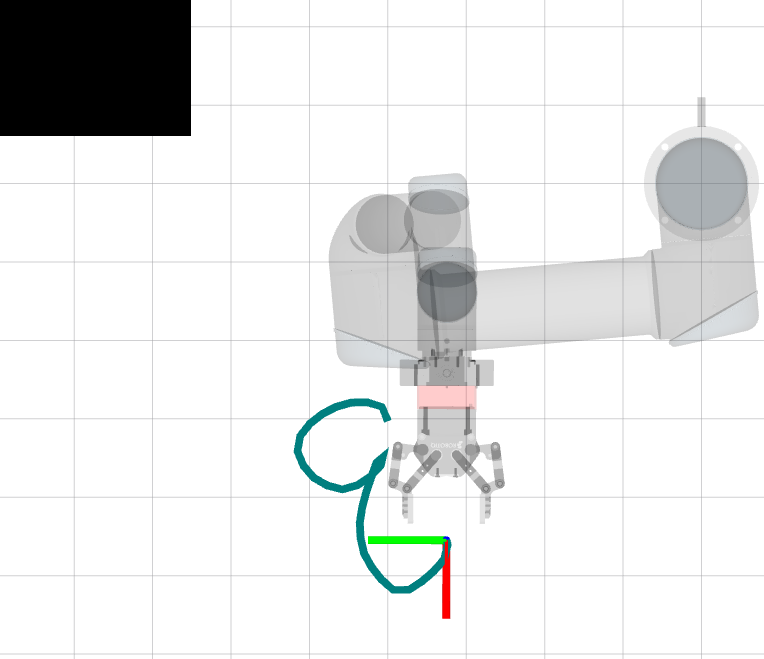}}
  \subfloat{\lettertrim{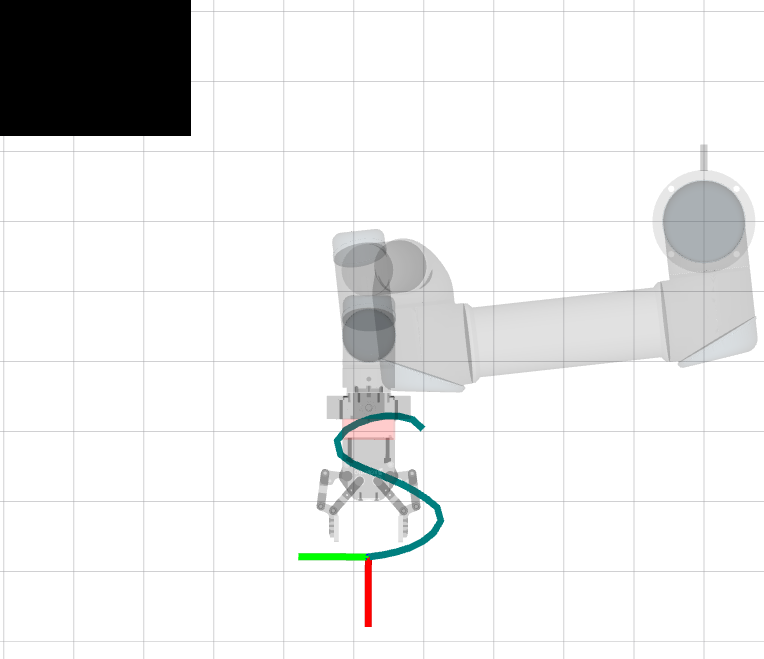}}
\caption{Examples of tracking human handwritten characters with a UR5 robot}
\label{fig: rviz simulations top view}
\end{figure}

%%%
\section{Model predictive control framework for tracking control}
\label{sec: mpc}
A general MPC formulation can be expressed as:
% \begin{equation}
\[
\begin{aligned}
  \V{\tilde{u}} = 
  & {\text{argmin}}
  & &
  \M{J}(\V{x}_0, \V{\tilde{u}})
  \\
  & \text{subject to}
  & & \V{x}_\text{i+1} = f(\V{x}_\text{i}, \V{u}_\text{i})
  ,
\end{aligned}
\]
% \label{eq: general mpc control problem}
% \end{equation}
where the resulting sequence of future control actions, 
$\V{\tilde{u}} = [\V{u}_\text{0} \hdots \V{u}_\text{N}]$ is obtained by optimizing 
the objective function, $\M{J}(\cdot)$, under the constraint of the system 
dynamics equation where $\V{x}_\text{i+1}$ is the resulting next state when the 
action, $\V{u}_\text{i}$ is applied while in state $\V{x}_\text{i}$.

The functions $\M{J}(\cdot)$ and/or 
$f(\cdot)$ can be fully or partially replaced by neural network models. This design 
choice leads to several different approaches. Here, we are using a neural network 
only as a part of $\M{J}(\cdot)$ and design it such that the neural 
network is a model to predict the future human motion.

For $f(\cdot)$, we assume that we can freely control the 
end-effector and that the motion is smooth so that the 
trajectory is differentiable three times.
Doing so, we can define the end-effector state as the Cartesian position, 
velocity and acceleration. We then use the jerk for control. For a single time step, 
$\timestep$ and a single degree of freedom (DOF), the equation for $f(\cdot)$ is 
linear such that:
% \begin{equation}
$
  \V{x}_\text{i+1} = \M{A}\V{x}_\text{i} + \M{B}\V{u}_\text{i}
  ,
$
%   \label{eq: one dof dynamics model}
% \end{equation}
where:
\[
  \V{x}
  =
  \begin{bmatrix}
    c \\
    \dot{c} \\
    \ddot{c}
  \end{bmatrix}
  ,
%   \hfill
  \V{u}
  = 
  \begin{bmatrix}
    \dddot{c}   
  \end{bmatrix}
  ,
%   \hfill
  \M{A} = 
  \begin{bmatrix}
    1	& \timestep  & \frac{\timestep^2}{2} \\
    0	& 1		   & \timestep \\
    0	& 0		   & 1
  \end{bmatrix}
  ,
%   \hfill
  \M{B} = 
  \begin{bmatrix}
    \frac{\timestep^3}{6} \\
    \frac{\timestep^2}{2}  \\
    \timestep
  \end{bmatrix}
%   .
\]
% \end{equation}
We can apply the same model independently for the three translations.
To obtain a vector of future states, 
$\V{\tilde{x}} = [\V{x}_\text{0} \hdots \V{x}_\text{N}]$,
we can recursively apply 
$f(\cdot)$
to get an 
arbitrarily long sequence of $N$ future states.
Doing so,
$\V{\tilde{x}}$ has length $9N$ (position, velocity, acceleration for 3DOF and N 
timesteps) while $\V{\tilde{u}}$ is a column vector, with length $3N$ (jerk for 
3DOF and N timesteps). A linear model for $f(\cdot)$ can still be written such that:
% \begin{equation}
$
  \V{\tilde{x}} = \M{\tilde{A}}\V{x}_\text{0} + \M{\tilde{B}}\V{\tilde{u}}
  ,
$
%   \label{eq: mpc model}
% \end{equation}
where $\V{x}_\text{0}$ is the initial state. The matrices 
$\M{\tilde{A}}, \M{\tilde{B}}$ are made from $\M{A}, \M{B}$ through a 
process known in MPC literature as \textit{condensing}.

For the objective function, we need to track the motion of the
character being written. This can be done by minimizing 
$\norm{\V{\tilde{x}}_\text{target} - \V{\tilde{x}}}$, the L2-norm of the state to 
a target state. $\V{\tilde{x}}_\text{target}$ requires future 
information so we need a predictive model. 
Here, we propose switching between two models: a 
conservative model, $\V{\tilde{x}_c}$, which predicts \textit{no motion}: just 
copying the last position with zero velocities and acceleration. 
This simple model has significant 
tracking error especially with quick motions but produces slower, more 
conservative motions since it is similar to only doing feedback control without 
prediction.
The other model is a Neural Network 
$g_{\boldsymbol{\theta}}(\V{x}_h)$ which takes as input a running history of the 
current state, $\V{x}_h$, and produces the prediction. This is explained in the next 
section. The last term of the objective is for smoothing out the control action. The 
final objective function is then built by adding gains $\M{G}_c, \M{G}_f$ and 
weights $\alpha, \beta$:
% \begin{equation}
\[
\begin{aligned}
  \M{J}(\V{x}_0, \V{\tilde{u}})
  =
  &
  (1-\alpha)
  \norm{\M{G}_c(\V{\tilde{x}_c} - \V{\tilde{x}})}^2
  +
  \\
  &
  \alpha
  \norm{\M{G}_f(g_{\boldsymbol{\theta}}(\V{x}_h) - \V{\tilde{x}})}^2
  +
  \beta
  \norm{\V{\tilde{u}}}^2
\end{aligned}
\]
% \label{eq: mpc control problemn objective}
% \end{equation}
The first two objectives are designed to 
achieve the same goal, so the weights are designed to be a homotopy with
$
  0 \leq \alpha \leq 1
$
.
Normally, only one of these objectives are active so
that 
$\alpha = 1$ or $\alpha = 0$.
% , otherwise if the targets of the two objectives 
% are different then these will be in conflict. 
However, when switching, a small 
transition period is needed where $\alpha$ is varied smoothly.
Meanwhile we only need the last term for smoothing/regularization so:
$
  \beta << 1
  .
$
The resulting control problem
can be solved quickly and efficiently as a 
quadratic programming (QP) problem.

%%%
\section{Human motion prediction with neural networks}
\label{sec: ml}
Human motion prediction with neural networks is also a topic of interest outside 
robot control, for example in~\cite{martinez:cvpr:2017, vinayavekhin:icpr:2018}.
The implicit assumption
is that there is an 
underlying motion pattern such that given a sufficiently long history of the current 
motion $\V{x}_h = [\V{x}_0 \hdots \V{x}_{t-1}]$,
we can predict $\V{x}_t = g_{\boldsymbol{\theta}}(\V{x}_h)$ by learning the 
parameters $\boldsymbol{\theta}$ of the neural network model $g$. For example, in 
our task when most of the letter is written, it should be clear which letter 
it is and this provides enough context to predict the future motion.
A well-known issue here is that the first few predictions will be bad since there is 
not enough history to provide a proper context yet. This is why we added the 
conservative model in our MPC and the functionality to switch between models.

The problem of human motion prediction is a well-studied subclass of 
sequence 
modeling where Recurrent Neural Networks (RNNs) have shown good results.
% In the current state-of-the-art,
The \textit{Long Short-Term 
Memory} (LSTM) model
~\cite{hochreiter:nc:1997} 
is the current standard for RNNs and used in benchmarks, for example 
in~\cite{martinez:cvpr:2017, vinayavekhin:icpr:2018}.
Although these RNN models have shown impressive results in several application 
areas, one concern here is the training time because all these models require back 
propagation through time. This is clearly not suited for online learning. At testing 
time, the forward pass is fast enough to be suitable for the 
robot control application we present. The disadvantage is that once the model 
is trained it has to be kept as it is.

% Apart from conventional RNNs, 
Another neural network model that is suitable for 
time-series prediction is the Dynamic Boltzmann 
Machine (DyBM) presented in~\cite{osogami:naturesr:2015}. It is an energy-based 
model designed 
for time-series prediction
with training speed considerations in mind so it does not use backpropagation 
through time.
It is also designed for online learning for edge devices.
Recently,~\cite{dasgupta:aaai:2017} compares a variation of the DyBM with 
the LSTM and the results are comparable in terms of prediction error.
The advantage is that the reported training time of the DyBM is $1/16$ of the LSTM. 
This is a significant advantage for our target applications.

Apart from the specific architecture of the neural network model, another design 
choice is the method for training which would dictate the function learned.

We are training 
the network to do a one-step prediction.
To produce the 
required $N$-steps prediction, we use the previous prediction result as 
the next input. A known issue of this technique is that the predictions 
will progressively worsen.
This does not affect the MPC since it
has a structure where later predictions have less weight 
in the optimization procedure.
An advantage of this technique is that $N$ can be arbitrarily set as the model is 
independent from it.
Lastly, after the $N$-steps prediction is created, the internal state of the LSTM 
and DyBM should be reset to just after the first prediction. This ensures continuity 
of the \textit{real} input sequence inside the memory of the NNs.

%----------------
\section{Results and discussion}
\label{sec: Results and discussion}
We evaluate our framework on the human handwriting dataset 
provided 
by~\cite{laviola:pami:2007}. The data is composed of the 
alphanumeric characters and basic math symbols written several times by 11 people. 
It is already divided into three sets: two training sets and a 
testing set. Here, we used only the ``\textit{training1}'' set consisting of 
$6590$ sequences for 
training. All the tests and validation are then done using the 
``\textit{testing}'' 
set which has $8136$ sequences. The data itself is composed of a series of positions 
in a 2-DOF coordinate 
system. As a normalization step, the series of positions are converted to 
velocities by finite differencing. The pen-up and pen-down events 
are removed such that there is a large computed velocity during this event.
At the end, $5$ zeros are appended to learn the concept of stopping after the 
writing stroke.
When training, the networks are reset before a new sequence is shown. 
Finally, we did not add any distinguishing mark for different characters and we used 
all the characters to train a single model. This is because we wanted the
neural networks to learn a general motion model which is suitable for all the 
character writing strokes.

% In the proposed method, one layer of
For this test, we used one layer of LSTM, with cell state of size $10$ and the 
$\tanh$ activation function.
% ~\cite{hochreiter:nc:1997} 
This is followed by a fully connected linear layer which produces the output.
The Mean Squared
Error (MSE) is used as a cost function for backpropagation. The model is
trained for $40$ epochs, with a batch size of 16 sequences which are zero-padded for 
uniformity.

As for the DyBM\footnote{https://github.com/ibm-research-tokyo/dybm}, 
we used the \textit{linear} version as the base with 
three different variations. Firstly, we trained it only offline with the training 
data. This serves as a comparison with the LSTM, which 
can only be trained offline for our application. Secondly, we allowed the DyBM to 
use the testing data for online learning. This is the normal usage of the DyBM. 
Finally, we added an echo 
state network (ESN)~\cite{jaeger:science:2004}, with size $50$ and leak parameter 
$0.7$, to the DyBM. This should enhance the 
non-linearities it can learn while still being fast enough for online learning.

%---
\subsection{Neural network inference results}
To serve as a baseline for evaluating the results, we used
the simplest sensible
prediction which is to assume that the velocity will remain constant. A similar 
model 
was used in~\cite{martinez:cvpr:2017} as a baseline for predicting human motion.
Table~\ref{table: training metrics} shows a summary of the results on the testing 
set. We are using 3 metrics: first the Mean Squared Error (MSE) over the whole 
validation set. Next, we do a Per-Sequence (PS) comparison. PS-B is the percentage 
of 
sequences having an MSE better than the baseline. PS-LSTM is similar but compared 
against the LSTM.
\begin{center}
  \resizebox{\columnwidth}{!}
  {
  \begin{tabular}{ | l | c | r | r |}
    \hline
    algorithm 	 & MSE 		& PS - B & PS - LSTM  \\ \hline
    baseline 	 & 3.0875 	& ---    & 33\%       \\ \hline
    LSTM 	 & 3.7132 	& 67\%   & ---	      \\ \hline
    DyBM offline & 3.2483	& 39\%   & 31\%       \\ \hline
    DyBM online  & 2.7151	& 79\%   & 39\%       \\ \hline
    DyBM online and ESN & \textbf{2.2715} & \textbf{90\%}& 42\%  \\ \hline
    \end{tabular}
    }
\label{table: training metrics}
\end{center}

We can see that for the mean squared error (MSE), the LSTM model and the DyBM 
trained only offline are both worse than the baseline.
However, the DyBMs with online learning are both better. The MSE here is just an 
indicator of the general model. To investigate further, we did per-sequence 
comparisons.
All the models except for DyBM with offline training are better than the baseline in 
more than $50\%$ of the 8136 validation sequences.
The results here are expected for the DyBMs but somewhat surprising for the LSTM 
which had a high overall MSE.
In checking this further, we observed that the sequences for the same symbols 
exhibit similar results. 
The LSTM performed worse in \textit{simpler, straigther} symbols such as ``v'', 
``1'', ``-'' but it was better in \textit{more complex, curvier} symbols 
such as ``p'', ``b'', ``0'' or those with discontinuities from parsing the 
pen-up-pen-down event like ``K''. Since the simple baseline should provide a 
good approximate for the simple symbols, it is better in these cases. 
Because the LSTM showed a good performance in the more difficult characters, this 
led to the comparison of PS-LSTM which is still per-sequence but against the LSTM. 
In this column, we see that the other methods overachieved LSTM only in less than 
$50\%$ of the sequences, although the online DyBMs are close at around $40\%$.

As a summary, the LSTM has 
learned a highly non-linear model which generalizes to different character strokes 
but at the cost of being much worse in simple character strokes leading 
to a high overall MSE. The DyBM trained only offline performs poorly across 
all metrics, but was not intended to be used in such manner. The online DyBM has 
learned a general model (high MSE, high PS-B). It is better than the LSTM for simple 
characters but worse for complex characters. The online DyBM with ESN is the best 
considering overall performance, but it is still slightly worse than the LSTM on 
complicated characters.

As for training speed,
the LSTM was trained with a batch size of 10 and took around 215 seconds per epoch, 
while the plain DyBM took around 43 seconds per epoch and with ESN around 54 seconds 
per epoch. Although 
not as high as for the dataset reported 
in~\cite{dasgupta:aaai:2017}, we see that it is still significantly faster.

\subsection{Results of the complete framework}
This subsection reports the results on testing the complete framework on simulations 
of a UR5 robot. 
Fig.~\ref{fig: rviz simulations top view} shows some results of the task.
For reference, the grid in Fig.~\ref{fig: rviz simulations top view} has a spacing of 
0.1 m.
For comparisons of how much the tracking error can be improved, we used a sequence 
for the letter K as a representative of the
results where the baseline performs poorly in terms of MSE. 
The sequence, taken from the validation set, is played online to represent the 
human writing the letter. The robot 
task is to try to write the letter together with the human at the exact same time. 
To control the robot, the MPC 
% of Eq.(\ref{eq: mpc control problem}) 
is used to 
generate the writing motion. This is then used as an end-effector command. Joint 
trajectory commands are obtained from this by 
using another QP for doing inverse kinematics, which handles the joint limits.

\begin{figure}
  \centering
  \includegraphics[width=\columnwidth]
  {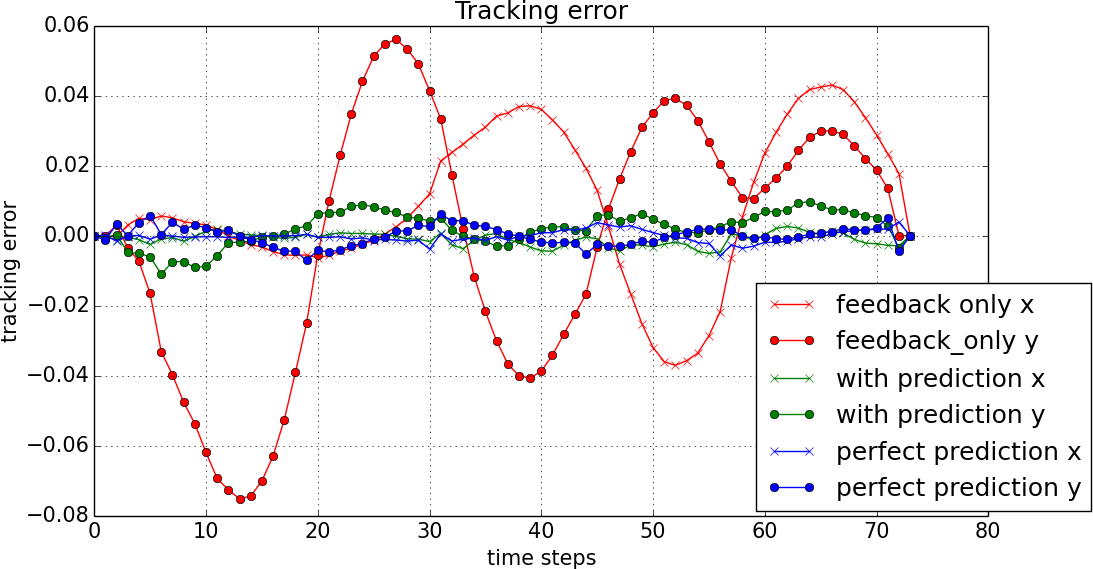}
  \caption{Tracking error (in meters) of the K letter. Red: MPC without 
prediction (feedback only). Green: Our method. Blue: MPC with a prefect 
prediction}
  \label{fig:results:combined tracking error comparison}
\end{figure}
Fig.~\ref{fig:results:combined tracking error comparison} shows a comparison of 
writing the same sequence in three different ways. First, only the feedback 
component was used to give a baseline. Secondly, we used one of the trained NN 
model's 
predictions while using the feedforward term all throughout. Lastly, a 
\textit{perfect prediction} can be done by using the test sequence in the 
feedforward term. Although this is practically impossible when the system runs 
online, it provides an ideal comparison point for the tests here.
We can see that the feedback-only case resulted in a tracking error going up to $7$ 
cm. The mean squared tracking error was about $0.0018 \text{m}^2$. In comparison, we 
can see a significant improvement by ``with prediction'' which used the LSTM with 
the preview horizon of length $10$ as a feedforward network for the MPC. Its mean 
squared tracking error was about $3.12\times10^{-5}\text{m}^2$.
This is an order of magnitude better than the feedback-only case.
Finally, we compare this result to a \textit{perfect prediction}, whose mean 
squared tracking error is about 
$1.01\times10^{-5}\text{m}^2$.
In this ideal prediction case, the error comes from a combination of 
the preview horizon (optimizing only on a limited time horizon instead of 
giving the full trajectory at once), the low-level robot motion controllers and 
the smoothing term of 
minimizing the jerk. The important point here is that using the NN for the 
feedforward term can result in tracking errors of the same order of magnitude as the 
\textit{perfect prediction} case.

\begin{figure}
  \centering
  \includegraphics[width=\columnwidth]
%   {K_letter_tracking_error_single_adaptive.png}
  {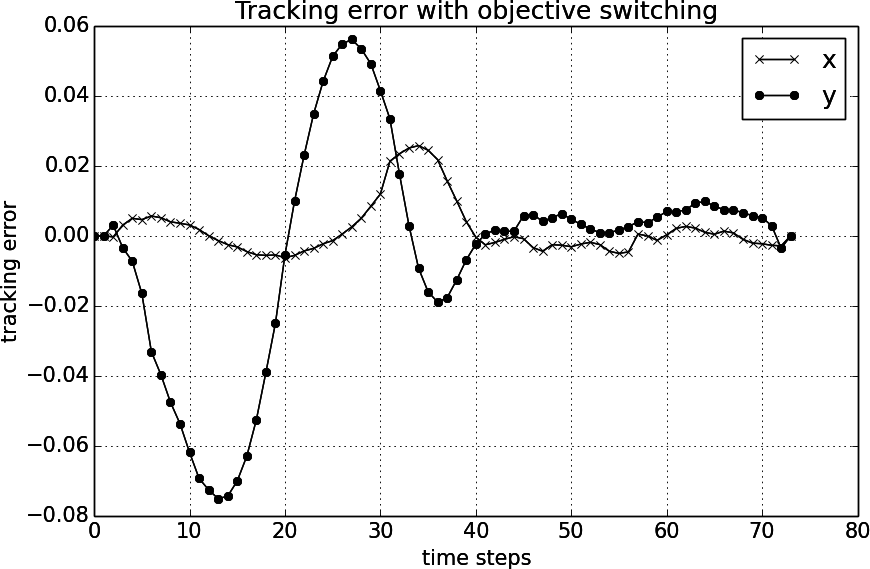}
  \caption{Tracking error (in meters) of the K letter switching 
from feedback-only into feedforward-only}
  \label{fig:results:tracking error adaptive}
\end{figure}

The final test is on using the weights to switch smoothly from feedback only to 
feedforward. The purpose of this test is to verify that there are no adverse effects 
due to the switching. The same sequence as those in
Fig.~\ref{fig:results:combined tracking error comparison} was used. The resulting 
tracking error is 
shown in Fig.~\ref{fig:results:tracking error adaptive}. The weight $\alpha$ 
was linearly decreased from $1$ to $0$ during time step $30$ until 
$40$. Fig.~\ref{fig:results:tracking error adaptive} shows no irregularity during 
this period where the error decreased as expected.

%%%
\section{Conclusion}
\label{sec: Conclusion}
In this paper, we presented a framework that can predict human motions by using 
different memory-based neural network models
and then effectively use these to produce an anticipatory
action by using an MPC. Furthermore, separate feedback and feedforward terms 
were designed to be able to cope with cases when the prediction is 
unreliable. Finally, we also demonstrated that it is possible to switch between 
the feedback and feedforward objectives seamlessly. The results show that  
the presented framework is an effective control strategy for human motion control
tracking tasks. Future works on using the same framework for various applications 
are planned.

\small
\bibliographystyle{IEEEtran}
\bibliography{2018-rsj-arxiv.bib}
\normalsize
\end{document}